\pdfoutput=1

\documentclass[11pt]{article}

\usepackage{emnlp2021}

\usepackage{times}
\usepackage{latexsym}

\usepackage[T1]{fontenc}

\usepackage[utf8]{inputenc}

\usepackage{microtype}
\usepackage{xcolor}
\usepackage{amsmath}
\usepackage{caption}
\usepackage{subcaption}
\usepackage{graphicx}
\usepackage{tabularx}
\usepackage{colortbl}
\usepackage[export]{adjustbox}

\title{A Benchmark Study of Contrastive Learning for Arabic Social Meaning}

\author{Md Tawkat Islam Khondaker$^{\dagger}$~~ { El Moatez Billah Nagoudi$^{\dagger}$}~~ { AbdelRahim Elmadany$^{\dagger}$}~~ \\ { \bf Muhammad Abdul-Mageed$^{\dagger}$}~~ { \bf Laks V.S. Lakshmanan}
\\\\ 
\normalsize $^{\dagger}$Deep Learning \& Natural Language Processing Group  \\
\normalsize The University of British Columbia\\
\\
\texttt{\{tawkat@cs.,laks@cs.,muhammad.mageed@\}ubc.ca}
}

\begin{document}
\maketitle
\begin{abstract}
Contrastive learning (CL) brought significant progress to various NLP tasks. Despite this progress, CL has not been applied to Arabic NLP to date. Nor is it clear how much benefits it could bring to particular classes of tasks such as those involved in Arabic social meaning (e.g., sentiment analysis, dialect identification, hate speech detection). In this work, we present a comprehensive benchmark study of state-of-the-art supervised CL methods on a wide array of Arabic social meaning tasks. Through extensive empirical analyses, we show that CL methods outperform vanilla finetuning on most tasks we consider. We also show that CL can be data efficient and quantify this efficiency. Overall, our work allows us to demonstrate the promise of CL methods, including in low-resource settings.
\end{abstract}

\section{Introduction}

Proliferation of social media resulted in unprecedented online user engagement. People around the world share their emotions, fears, hopes, opinions, etc. online on a daily basis (\citealt{farzindar,zhang2022improving}) on platforms such as Facebook and Twitter. Hence, these platforms offer excellent resources for social meaning tasks such as emotion recognition (\citealt{mageed2017emonet,saif_emotion}), irony detection (\citealt{van_hee}), sarcasm detection (\citealt{bamman}), hate speech identification (\citealt{waseem_2016}), stance identification (\citealt{saif_stance}), among others. While the majority of previous social meaning studies were carried out on English, a fast-growing number of investigations focus on other languages. In this paper, we focus on Arabic.

Several works have been conducted on different Arabic social meaning tasks. Some of these focus on Modern Standard Arabic (MSA) (\citealt{muhammad_2011, mageed2012samar}), while others take Arabic dialects as their target (\citealt{elsahar2015building,al2015deep}). While many works have focused on sentiment analysis, e.g., ~\cite{mageed2012samar,nabil2015astd,elsahar2015building,al2015deep,al2018arabic,al2019using,al2019comprehensive,farha2019mazajak} and dialect identification~\cite{elfardy2013sentence,zaidan2011arabic,zaidan2014arabic,cotterell2014multi,zhang2019no,Bouamor:2018:madar,mageed2020microdialect,mageed-etal-2020-nadi,abdul-mageed-etal-2021-nadi}, others focused on detection of user demographics such as age and gender (\citealt{zaghouani2018arap, rangel}), irony detection (\citealt{karoui, idat2019}), and emotion analysis (\citealt{muhammad_emotion, alhuzali2018enabling}). Our interest in the current work is improving Arabic social meaning through representation learning.


\begin{figure}[t]
    \centering
    \includegraphics[width=\columnwidth]{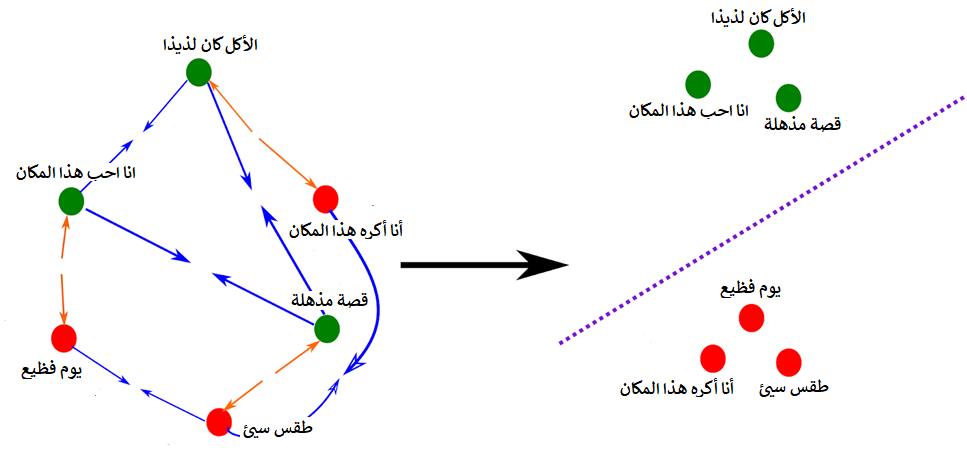}
    \caption{Visual illustration of how supervised contrastive learning works. Representations from the same class are \textcolor{blue}{\textit{pulled}} close to each other while representations from the different classes are \textcolor{orange}{\textit{pushed}} further apart.}
    \label{fig:demo_cl}
\end{figure}


In spite of recent progress in representation learning, most work in Arabic social meaning mostly focuses on finetuning language models such as AraT5~\cite{nagoudi-etal-2022-arat5}, CamelBERT~\cite{inoue-etal-2021-interplay}, MARBERT~\cite{marbert}, QARIB~\cite{abdelali2021pre}, among others. In particular, Arabic social media processing has to date ignored the emerging sub-area of contrastive learning (CL) (\citealt{hadsell}). Given a labeled dataset, CL \citep{scl} attempts to pull representations of the same class close to each other while pushing representations of different classes further apart (Figure \ref{fig:demo_cl}). In this work, we investigate five different supervised contrastive learning methods in the context of Arabic social meaning. To the best of our knowledge, this is the first work that provides a comprehensive study of supervised contrastive learning on a wide range of Arabic social meanings. We show that performance of CL methods can be task-dependent. We attempt to explain this performance from the perspective of task specificity (i.e., how fine-grained the labels of a given task are). We also show that contrastive learning methods generally perform better than vanilla finetuning based on cross entropy (CE). Through an extensive experimental study, we also demonstrate that CL methods outperform CE finetuning under resource-limited constraints. Our work allows us to demonstrate the promise of CL methods in general, and in low-resource settings in particular.

To summarize, we offer the following contributions:

\begin{enumerate}
    \item We study a comprehensive set of supervised CL methods for a wide range of Arabic social meaning tasks, including abusive language and hate speech detection, emotion and sentiment analysis, and identification of demographic attributes (e.g. age, gender).
    
    \item We show that CL-based methods outperform generic CE-based vanilla finetuning for most of the tasks. To the best of our knowledge, this is the first work that provides an extensive study of supervised CL on Arabic social meaning.
    
    \item We empirically find that improvements CL methods result in are task-specific and attempt to understand this finding in the context of the different tasks we consider with regard to their label granularity.
    
    \item We demonstrate that CL methods can achieve better performance under limited data constraints, emphasizing and quantifying how well these can work for low-resource settings.
    
\end{enumerate}

\section{Related Works}

\subsection{Arabic Social Meaning}\label{sec:lit_sm}
We use the term \textit{social meaning} (SM) to refer to meaning arising in real-world communication in social media~\cite{thomas2014meaning,zhang2022decay}. SM covers tasks such as sentiment analysis~\cite{mageed2012samar,abu-farha-etal-2021-overview, saleh2022heterogeneous, alali2022multitasking}, emotion recognition~\cite{alhuzali2018enabling, mubarak2022emojis, abu2022multi, mansy2022ensemble}, age and gender identification~\cite{abdul2020aranet,abbes2020daict,mubarak2022arabgend, mansour2022towards}, hate-speech and offensive language detection~\cite{elmadany2020leveraging,mubarak-etal-2020-overview, mubarako-2022-verview, husain2022investigating}, and sarcasm detection~\cite{farha2020arabic, wafa2022sarcasm, abdullah2022sarcasmdet}.\\

Most of the recent studies are transformers-based. They directly finetune pre-trained models such as mBERT \cite{devlin2018bert}, MARBERT~\cite{marbert}, and AraT5~\cite{nagoudi-etal-2022-arat5} on SM datasets like~\cite{abdul2020aranet, alshehri2020understanding,abuzayed2021sarcasm, nessiricompass},  using data augmentation~\cite{elmadani_osact4}, ensampling~\cite{mansy2022ensemble, alzu2022aixplain}, and multi-tasks~\cite{mageed2020microdialect,shapiro2022alexu, alkhamissi2022meta}. However, to the best of our knowledge, there is no published research studying CL on Arabic language understanding in general nor social meaning processing in paticular.


\subsection{Contrastive Learning}

CL aims to learn effective embedding by pulling semantically close neighbors together while pushing apart non-neighbors (\citealt{hadsell}). CL employs a CL-based similarity objective to learn the embedding representation in the hyperspace~\cite{chen_2017,henderson_2017}. In computer vision,~\citet{simclr} propose a framework for contrastive learning of visual representations without specialized architectures or a memory bank.~\citet{scl} shows that supervised contrastive loss can outperform CL loss on ImageNet~\cite{imagenet}. In NLP, similar methods have been explored in the context of sentence representation learning~\cite{karpukhin,gillick,logeswaran,zhang2022infodcl}. Among the most notable works is~\citet{simcse} who propose unsupervised CL framework, \textit{SimCSE}, that predicts input sentence itself by augmenting it with dropout as noise.

Recent works have been studying CL extensively for improving both semantic text similarity (STS) and text classification tasks (\citealt{cocolm, coda, carl, contrastive_tension}).~\citet{cert} propose back-translation as a source of positive pair for NLU tasks.~\citet{scd} argue that feature decorrelation between high and low dropout projected representations improves STS tasks.~\citet{dclr} design an instance weighting method to penalize false negatives and generate noise-based negatives to guarantee the uniformity of the representation space.~\citet{tacl} propose a token-aware CL method by contrasting the token from the same sequence to improve the uniformity in the embedding space. We now formally introduce these CL methods and how we employ them in our work.

\begin{figure*}[t]
     \centering
     \begin{subfigure}[b]{0.3\textwidth}
         \centering
         \includegraphics[width=\textwidth]{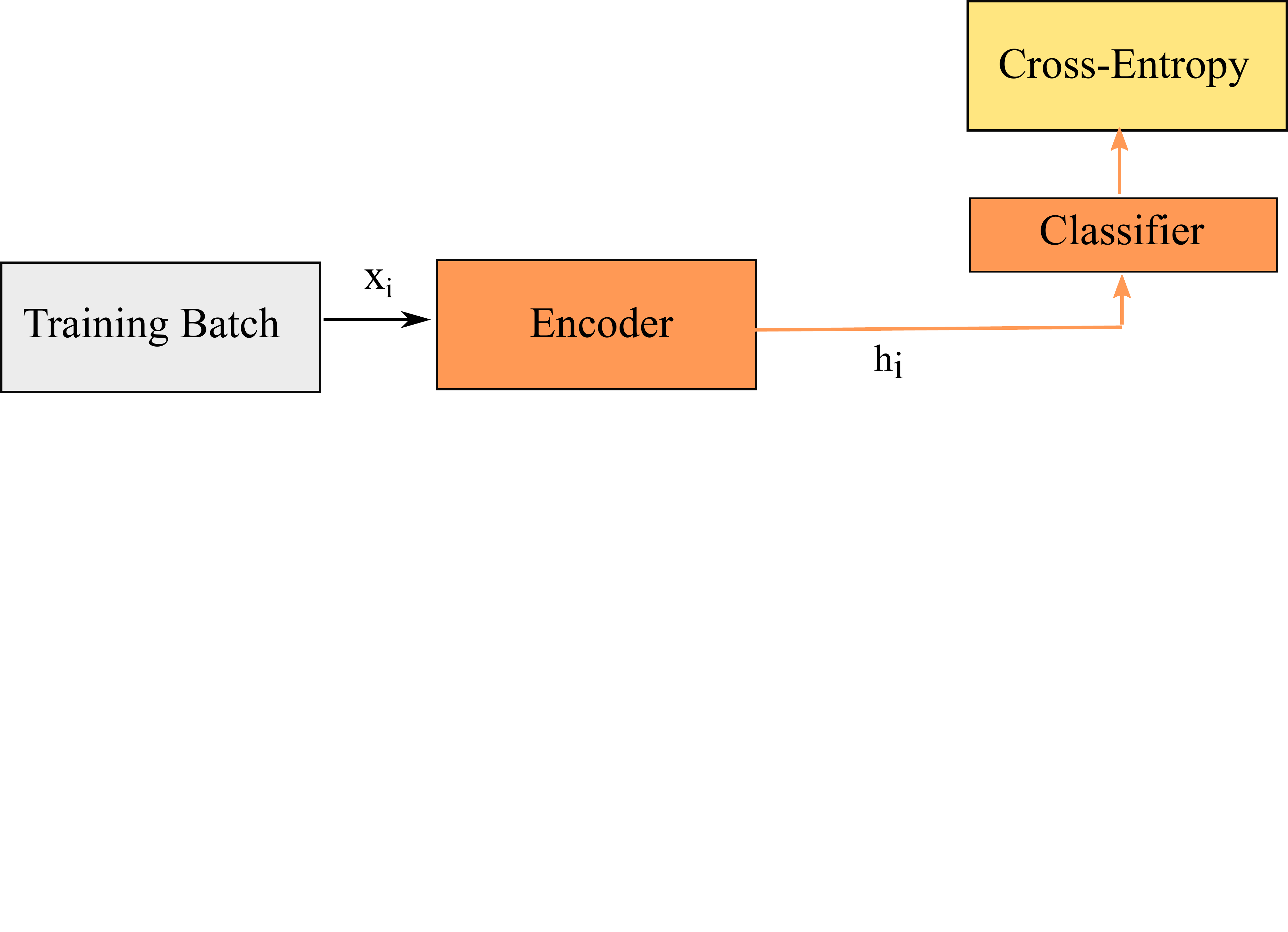}
         \caption{CE}
         \label{fig:ce}
     \end{subfigure}
     \hfill
     \begin{subfigure}[b]{0.3\textwidth}
         \centering
         \includegraphics[width=\textwidth]{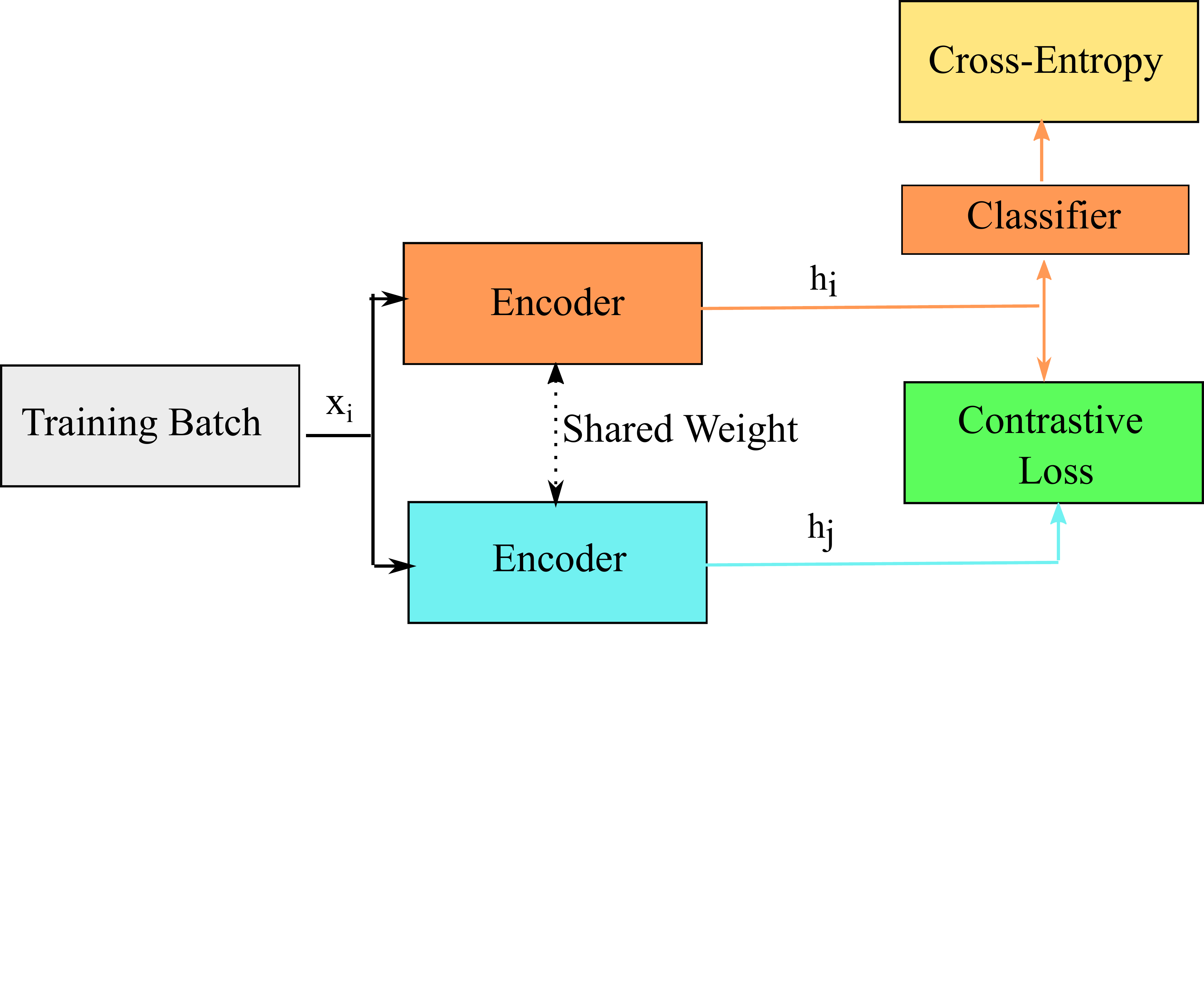}
         \caption{SCL}
         \label{fig:scl}
     \end{subfigure}
     \hfill
     \begin{subfigure}[b]{0.3\textwidth}
         \centering
         \includegraphics[width=\textwidth]{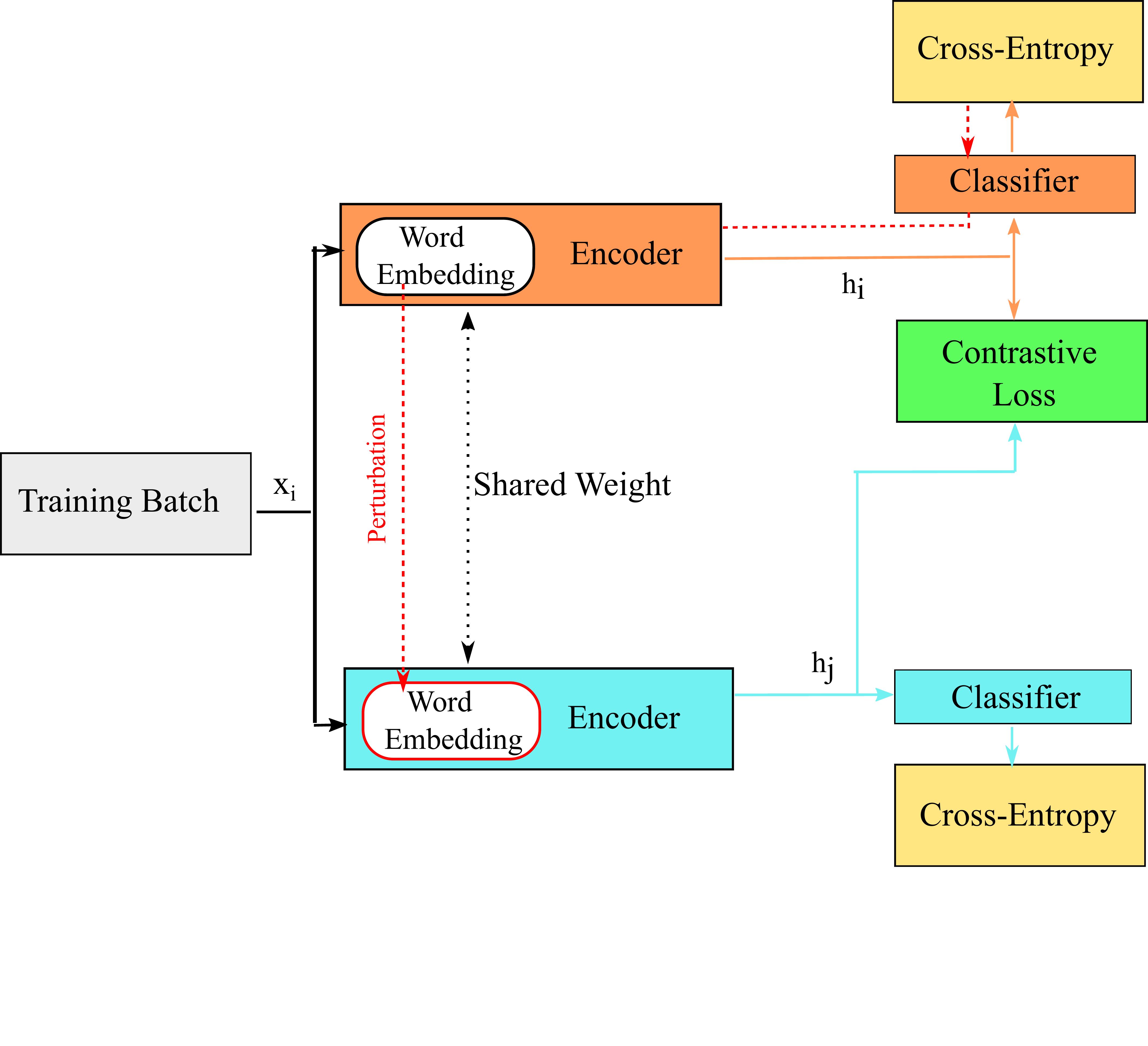}
         \caption{CAT}
         \label{fig:cat}
     \end{subfigure}
    \hfill
     \begin{subfigure}[b]{0.3\textwidth}
         \centering
         \includegraphics[width=\textwidth]{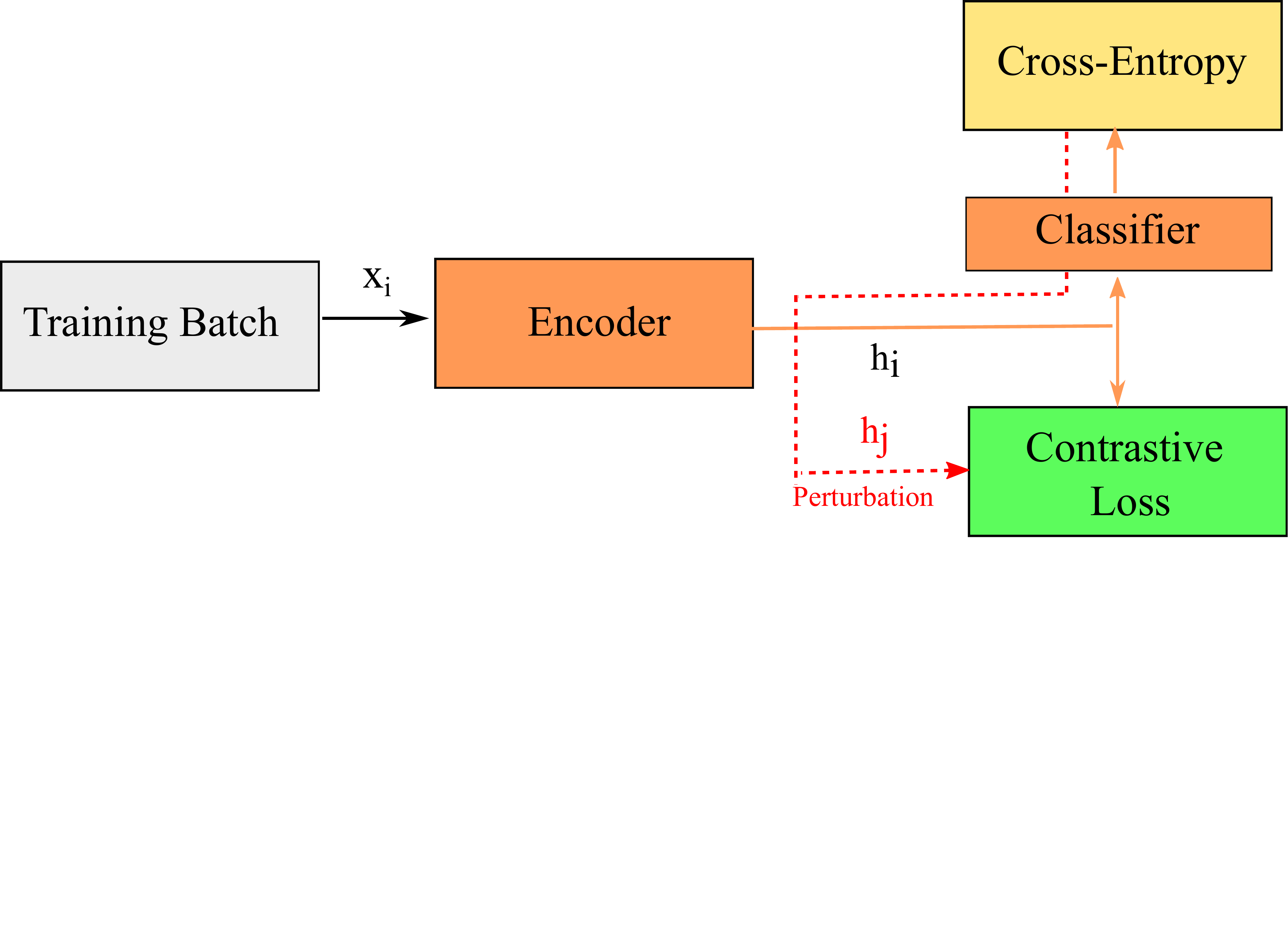}
         \caption{TACT}
         \label{fig:tact}
     \end{subfigure}
    \hfill
     \begin{subfigure}[b]{0.3\textwidth}
         \centering
         \includegraphics[width=\textwidth]{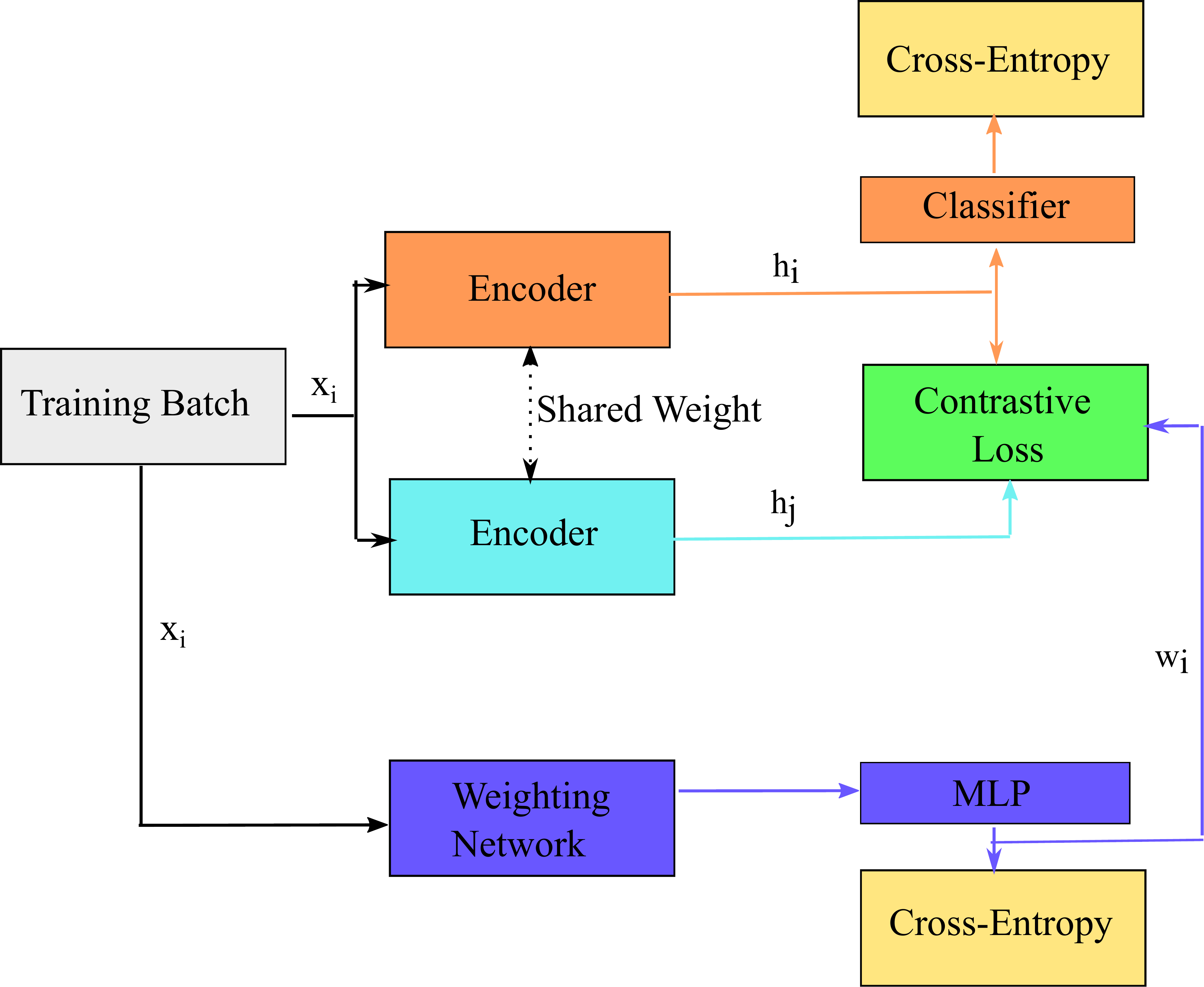}
         \caption{LCL}
         \label{fig:lcl}
     \end{subfigure}
    \hfill
     \begin{subfigure}[b]{0.3\textwidth}
         \centering
         \includegraphics[width=\textwidth]{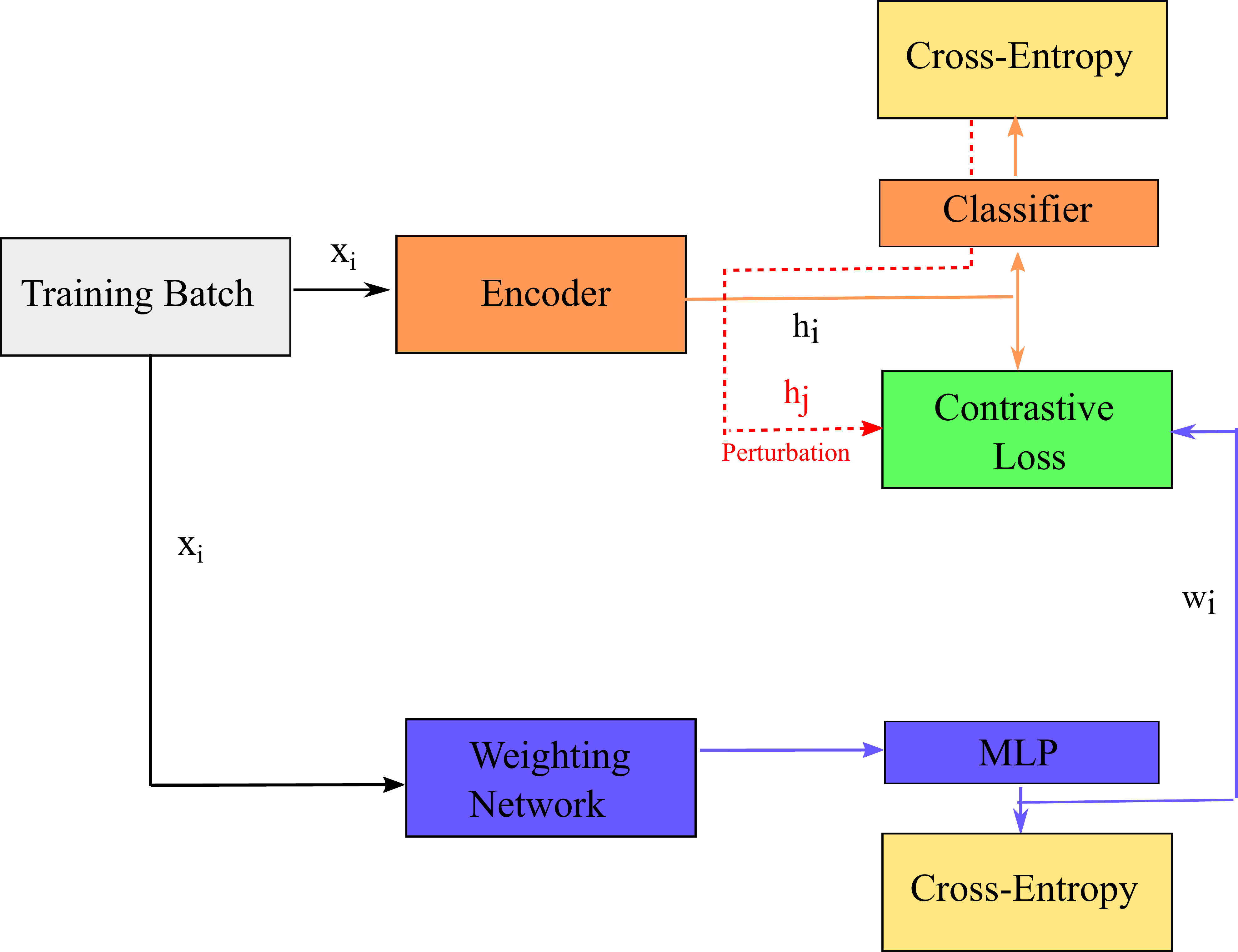}
         \caption{TLCL}
         \label{fig:tlcl}
     \end{subfigure}
        \caption{Illustration of supervised contrastive learning methods used in this work.}
        \label{fig:models}
\end{figure*}


\section{Methods}

Given a set of training examples $\{x_i, y_i\}_{i=1,...,N}$ and an encoder based on a pre-trained language model (PLM), \textit{f} outputs contextualized token representation of $x_i$,

\begin{equation}
    \mathit{H} = \{\;h_{[CLS]},\; h_1,\; h_2,\; ...,\; h_{[SEP]}\;\}
\end{equation}

\noindent Where \textbf{\textit{H}} is the hidden representation of the final layer of the encoder.

The standard practice of finetuning PLMs passes the pooled representation $h_{[CLS]}$ of \texttt{[CLS]} to a softmax classifier to obtain the probability distribution for the set of classes \textbf{C} (Figure \ref{fig:ce}).
\begin{equation}
    p(y_c|h_{[CLS]}) = softmax\; (\mathbf{W}h_{[CLS]}); \,\,\, c \in \mathbf{C}
\end{equation}
Where $\mathbf{W} \in \mathcal{R}^{d_C\; \times\; d_h}$ are trainable parameters and $d_h$ is hidden dimension. The model is trained with the objective of minimizing cross-entropy (CE) loss,

\begin{equation}
    \mathcal{L}_{CE} = -\frac{1}{N}\sum^N_{i=1}\sum^C_{c=1}y_{i,c} \; log(p(y_{i,c} |  h_{i_{[CLS]}}))\footnote{$h_{i_{[CLS]}}$ and $h_i$ are used interchangeably in the rest of the paper.}
\end{equation}

\subsection{Supervised Contrastive Loss (SCL)}
\label{sec:scl}

The objective of supervised contrastive loss (\citealt{scl}) is to pull the representations of the same class close to each other while pushing the representations of different classes further apart. Following \citet{simcse}, we adopt dropout-based data augmentation where for each representation $h_i$, we produce an equivalent dropout-based representation $h_j$ and consider $h_j$ as having the same label as $h_i$ (Figure \ref{fig:scl}).
The model attempts to minimize NTXent loss~\citep{simclr}. The purpose of NTXent loss is to take each in-batch representation as an anchor and minimize the distance between the anchor($h_i$) and the representations from the same class ($P_i$) while maximizing the distance between the anchor and the representation from different classes,

\small
\begin{equation}
\label{eq:scl}
    \mathcal{L}_{NTX} = \sum_{i=1}^{2N}\;\frac{-1}{P_i}\sum_{j \in P_i}\;\log\frac{e^{sim(h_i, h_j)/\tau}}{\sum_{k=1}^{2N}\;1_{i\neq k}e^{sim(h_i, h_k)/\tau}}
\end{equation}
\normalsize

\noindent Where $\tau$ is used to regulate the temperature. The final loss for SCL is
$$
\mathcal{L}_{SCL} = (1-\lambda)\mathcal{L}_{CE} + \lambda\mathcal{L}_{NTX}
$$

\subsection{Contrastive Adversarial Training (CAT)}

Instead of dropout-based augmentation,~\citet{cat} propose to generate adversarial examples applying \textit{fast gradient sign method} (FGSM) \citep{goodfellow_2015}. Formally, \textit{FGSM} attempts to maximize $\mathcal{L}_{CE}$ by adding a small perturbation $r$ bounded by $\epsilon$,

\begin{equation}
\begin{aligned}[b]
    max\mathcal{L}_{CE} = arg\max_r \mathcal{L}(f(x_i + r, y_i) \\s.t.\; ||r|| < \epsilon, \:\:\: \epsilon>0
\end{aligned}
\end{equation}

\noindent\citet{goodfellow_2015} approximate the perturbation $r$ with a linear approximation around $x_i$ and an \textit{L2} norm constraint. However,~\citet{cat} propose to approximate $r$ around the word embedding matrix $V \in \mathcal{R}^{d_V\; \times\; d_h}$ (Figure \ref{fig:cat}), where $d_V$ is the vocabulary size. Hence, the adversarial perturbation is computed as,

\begin{equation}
    r = -\epsilon \frac{\nabla_{V}\mathcal{L}(f(x_i, y_i)}{{||\nabla_{V}\mathcal{L}(f(x_i, y_i)||}_2}
\end{equation}

\noindent After receiving $x_i$, the perturbed encoder $f^{V+r}$ outputs \texttt{[CLS]} representation $h_j$, which is treated as the positive pair of $h_i$. Both $h_i$ and $h_j$ are passed through a non-linear projection layer and the resulting representations are used to train the model with  InfoNCE loss~\citep{infonce}.

\begin{equation}
    z_i = \mathbf{W_2}ReLU(\mathbf{W_1}h_i)
\end{equation}
\begin{equation}
    z_j = \mathbf{W_2}ReLU(\mathbf{W_1}h_j)
\end{equation}

\begin{equation}
\label{eq:infonce}
    \mathcal{L}_{InfoNCE} = -\log\frac{e^{sim(z_i, z_j)/\tau}}{\sum_{k=1}^{2N}\;1_{i\neq k}e^{sim(z_i, z_k)/\tau}}
\end{equation}

The final loss is calculated as,
$$
\mathcal{L}_{CAT} = \frac{1-\lambda}{2}(\mathcal{L}_{CE} + \mathcal{L}^{V+r}_{CE}) + \lambda\mathcal{L}_{InfoNCE}
$$

\subsection{Token-level Adversarial Contrastive Training (TACT)}
\label{sec:tact}

We also study a variant of CAT where instead of perturbing the word embedding matrix $V$, we directly perturb the token representations $h_i$ (Figure \ref{fig:tact}),

\begin{equation}
\label{eq:tact_r}
    r = -\epsilon \frac{\nabla_{h_i}\mathcal{L}(f(x_i, y_i)}{{||\nabla_{h_i}\mathcal{L}(f(x_i, y_i)||}_2}
\end{equation}
\begin{equation}
\label{eq:tact_h}
    h_j = h_i + r
\end{equation}

\noindent Similar to CAT, we pass $h_i$ and $h_j$ through a non-linear projection layer and use the obtained representations to train the model to minimize InfoNCE loss (Eq. \ref{eq:infonce}). We compute the final loss as,
\begin{equation}
    \mathcal{L}_{CAT} = \frac{1-\lambda}{2}(\mathcal{L}_{CE} + \mathcal{L}^{h+r}_{CE}) + \lambda\mathcal{L}_{InfoNCE}
\end{equation}

\subsection{Label-aware Contrastive Loss (LCL)}
\citet{lcl} propose to adapt contrastive loss for fine-grained
classification tasks by incorporating inter-label relationships. The authors propose an additional weighting network (Figure \ref{fig:lcl}) to encode the inter-label relationships. First, both the encoder and the weighting network are optimised using cross-entropy loss ($\mathcal{L}_{CE}$), $\mathcal{L}_E$, and $\mathcal{L}_w$, respectively. The prediction probabilities obtained from the softmax layer of the weighting network are used to compute the confidence of the current sample for a given class $c$,

\begin{equation}
    \mathbf{w}_{i,c} = \frac{e^{h_{i,c}}}{\sum_{k=1}^{C}\;e^{h_{i,k}}} 
\end{equation}

\noindent These weights are then used to train the model with NTXent loss.

\begin{equation}
\label{eq:lcl_ntx}
    \mathcal{L}_{i} = \sum_{j \in P_i}\log\frac{w_{i, y_i} \cdot e^{sim(h_i, h_j)/\tau}}{\sum_{k=1}^{2N}\;1_{i\neq k}w_{i, y_k}\cdot e^{sim(h_i, h_k)/\tau}}
\end{equation}
\begin{equation}
    \mathcal{L}_f = \sum_{i=1}^{2N}\;\frac{-\mathcal{L}_i}{P_i}
\end{equation}

\begin{table*}[t]
\centering
\setlength{\tabcolsep}{10pt}
\begin{tabular}{llllc}
\hline
\textbf{Dataset}   & \textbf{Train} & \textbf{Dev} & \textbf{Test} & \textbf{No. of Classes} \\ \hline
Abusive                 & 4,677                 & 584                & 585                 & 3                      \\
Adult                   & 33,690                & 5,000               & 5000                & 2                      \\
Age                     & 5,000                 & 5,000               & 5,000                & 3                      \\
AraNeT\textsubscript{emo}                 & 50,000                & 910                & 941                 & 8                      \\
Dangerous               & 3,474                 & 615                & 663                 & 2                      \\
Dialect at BinaryLevel  & 50,000                & 5,000               & 5,000                & 2                      \\
Dialect at CountryLevel & 50,000                & 5,000               & 5,000                & 21                     \\
Dialect at RegionLevel  & 38,271                & 4,450               & 5000                & 4                      \\

Gender                  & 50,000                & 5,000               & 5,000                & 2                      \\
Hate Speech             & 6,839                 & 1,000               & 2,000                & 2                      \\
Irony                   & 3,621                 & 403                & 805                 & 2                      \\
Offensive               & 6,839                 & 1,000               & 2,000                & 2                      \\
Sarcasm                 & 7,593                 & 844                & 2,110                & 2                      \\
SemEval\textsubscript{emo}    & 3,376                 & 661                & 1,563                & 4                      \\
Sentiment Analysis      & 49,301                & 4,443               & 4,933                & 3                      \\ \hline
\end{tabular}
\caption{
\label{table:datasets}
Statistics of datasets used in our experiments.
}
\end{table*}

\noindent Similar to Section \ref{sec:scl}, we use dropout-based data augmentation. Given a confusable sample, the weighting network will assign higher scores for the classes that are more closely associated with the sample. Incorporating these high values back into the denominator of NTXent will steer the encoder toward finding more distinguishing patterns to differentiate between confusable samples. The final LCL loss is computed as follows:

\begin{equation}
    \mathcal{L}_{LCL} = (1 - \lambda)(\mathcal{L}_E + \mathcal{L}_w) + \lambda\mathcal{L}_f
\end{equation}

\subsection{Token Adversarial LCL (TLCL)}

Instead of dropout-oriented representation as an augmentation, we experiment with token adversarial representation for LCL (Figure \ref{fig:tlcl}) described in Section \ref{sec:tact}. First, we compute the adversarial representation $h_j$ using Eq. \ref{eq:tact_r} and Eq. \ref{eq:tact_h}. Then, we compute NTXent loss (Eq. \ref{eq:lcl_ntx}) for LCL to obtain the final token adversarial LCL loss, $\mathcal{L}_{TLCL}$. We now describe our datasets.

\section{Datasets}

In this section, we present the Arabic social meaning tasks and datasets used in our study. A summary of the datasets is presented in Table \ref{table:datasets}.

\noindent \textbf{Abusive and Adult Content.} For the abusive and adult content detection tasks, we use  datasets from~\citet{mubarak-etal-2017-abusive} and \citet{mubarak2021adult}. These datasets consist of $1.1$k and $43$k tweets, respectively. For these datasets, the goal is to classify an Arabic tweet into one of the two classes in the set, i.e., \textit{\{obscene, clean\}} for the abusive task, and \textit{\{adult, not-adult\}} for the adult content detection task. 

\noindent \textbf{Age and Gender.}
For both tasks, we use the \textit{Arap-Tweet} dataset~\cite{zaghouani2018arap} which consists of \textit{1.3M, 160k, 160k}  for the Train, Dev, and Test respecctively. The dataset covers $11$ Arab regions.~\newcite{zaghouani2018arap}  assign age group labels from the set \textit{\{under-25, 25-to-34, above-35\}} and gender from the set \textit{\{male, female\}}.

\noindent \textbf{Dangerous.} We use the dangerous speech dataset from ~\newcite{alshehri2020understanding}. This dataset consists of $4,445$ manually annotated tweets labelled as either \textit{safe} or  \textit{dangerous}.

\noindent \textbf{Dialect Identification:} Six datasets are used for this task: ArSarcasm\textsubscript{Dia}~\cite{farha2020arabic},  the Arabic Online Commentary (AOC)~\cite{zaidan2014arabic}, NADI-2020~\cite{mageed-etal-2020-nadi}, MADAR~\cite{bouamor2019madar},  QADI~\cite{abdelali2020arabic}, and Habibi~\cite{el2020habibi}. The  dialect identification task involves three  dialect classification levels:  (1) Binary-level (\textit{MSA} vs. \textit{DIA}), (2) Region-level  ($4$  \textit{regions}), and  (3) Country-level  ($21$ \textit{countries}). 

\noindent \textbf{Emotion.} For this task, we use two datasets:  \textit{AraNeT\textsubscript{emo}} and \textit{SemEval\textsubscript{emo}}. The first one is proposed by~\citet{abdul2020aranet}. The dataset consists of $192$K tweets labeled with the eight emotion classes from the set \{\textit{anger, anticipation, disgust, fear, joy, sadness, surprise, trust}\}. \textit{SemEval\textsubscript{emo}} \citep{saif_emotion} consists of $5,603$ tweets labeled with four emotions from the set  \{\textit{anger, fear, joy, sadness}\}.

\noindent \textbf{Offensive Language and Hate Speech}. We use the dataset released by~\newcite{mubarak-etal-2020-overview} during an offensive and hate speech shared task.\footnote{\href{http://edinburghnlp.inf.ed.ac.uk/workshops/OSACT4/}{http://edinburghnlp.inf.ed.ac.uk/workshops/OSACT4/}} This dataset consists of $10k$  manually annotated tweets with four tags \textit{\{offensive, not-offensive, hate, not-hate}

\noindent \textbf{Irony}. We use the irony identification dataset for Arabic tweets (IDAT) developed by~\citet{idat2019}. This dataset contains $5,030$ MSA and dialectal tweets. It is labeled  with \textit{ironic} and \textit{non-ironic} tags.

\noindent \textbf{Sarcasm}. We use the \textit{ArSarcasm} dataset released by~\cite{farha2020arabic}. \textit{ArSarcasm} contains $10,547$ tweets. The tweets are labeled with \textit{sarcasm} and \textit{not-sarcasm} tags.

\noindent \textbf{Sentiment Analysis} This task includes $19$ sentiment datasets.  We merge the $17$ datasets benchmarked  by~\citet{marbert} with two new datasets:  Arabizi sentiment analysis dataset \cite{Chayma2020} and AraCust~\cite{almuqren2021aracust}, a Saudi Telecom Tweets corpus for sentiment analysis. The data contains \textit{190k, 6.5k, 44.2k} samples for Train, Dev and Test. The dataset is labeled with three tags from the set  \{\textit{positive}, \textit{negative}, \textit{neutral}\}.

\begin{table*}[]
\centering
\setlength{\tabcolsep}{15pt}
\begin{tabular}{@{}lcccccc@{}}
\hline
                        & {\bf CE}                            & {\bf SCL}                           & {\bf CAT}                           & {\bf TACT}                          & {\bf LCL}                           & {\bf TLCL}                          \\ \hline
Abusive                 & 77.15                         & 78.09                         & 76.48                         & 75.69                         & \textbf{78.32} & 75.26                         \\
Adult                   & 88.16                         & \textbf{89.50}  & 86.54                         & 89.13                         & 88.85                         & 89.48                         \\
Age                     & 44.22                         & 45.12                         & 42.28                         & \textbf{46.45} & 45.90                          & 43.20                          \\
AraNeT\textsubscript{emo}                 & 62.47                         & 61.49                         & 59.31                         & 57.99                         & 62.56                         & \textbf{64.13} \\
Dangerous               & 61.44                         & 63.76                         & 67.83                         & 66.00                            & 65.76                         & \textbf{69.28} \\
Dialect at BinaryLevel  & 85.71                         & 85.63                         & \textbf{86.67} & 84.98                         & 85.79                         & 81.84                         \\
Dialect at CountryLevel & 32.84                         & \textbf{33.63} & 33.24                         & 32.69                         & 33.62                         & 31.34                         \\
Dialect at RegionLevel  & 65.29                         & 64.78                         & \textbf{65.54} & 64.56                         & 62.92                         & 64.92                         \\

Gender                  & 62.23                         & 63.56                         & 65.58                         & 65.77                         & \textbf{65.90}  & 65.14                         \\
Hate Speech             & 80.91                         & 80.00                            & 71.06                         & \textbf{82.62} & 81.00                            & 75.26                         \\
Irony                   & \textbf{84.75} & 84.30                          & 84.72                         & 84.18                         & 84.29                         & 83.43                         \\
Offensive               & 90.43                         & 89.92                         & \textbf{91.37} & 91.23                         & 90.41                         & 88.84                         \\
Sarcasm                 & 70.67                         & 71.09                         & 72.09                         & 74.14                         & \textbf{75.32} & 69.40                          \\
\textit{SemEval\textsubscript{emo}}      & 79.25                         & 77.22                         & 77.08                         & 77.85                         & \textbf{80.61} & 78.59                         \\
Sentiment Analysis      & \textbf{77.69} & 77.32                         & 76.89                         & 76.68                         & 75.61                         & 74.82                         \\ \hline
Avg.                     & 70.88                         & 71.03                         & 70.45                         & 71.33                         & \textbf{71.79} & 70.33                         \\ \hline
\end{tabular}
\caption{
\label{table:model_performance}
Macro F1-score of the models on Arabic social media datasets. Here, \emph{CE} = Cross-Entropy; \emph{SCL} = Supervised Contrastive Learning; \emph{CAT} = Contrastive Adversarial Training; \emph{TACT} = Token-level Adversarial Contrastive Training; \emph{LCL} = Label-aware Contrastive Loss; \emph{TLCL} = Token Adversarial LCL.
}
\end{table*}

\section{Experimental Setup}

We implement all the methods using MARBERT~\citep{marbert} (\texttt{UBC-NLP/MARBERT}) from
HuggingFace's Transformers library \citep{wolf_huggingface}, as the backbone architecture. We use MARBERT as it is reported to achieve SOTA on a wide range of Arabic language understanding tasks in~\citet{marbert}. Our methods, however, can be applied to any other model. We use the same hyperparameters for all the methods to ensure fair comparisons. We set the maximum sequence length to $128$ and use a batch size of $16$ to train the models using Adam optimizer with a learning rate $5e-5$. The initial number of training epochs is set to $25$ with an early stopping threshold of $5$. For CL-based models, we set $\lambda$ to $0.5$ and $\tau$ to $0.3$. For all the experiments, we consider the checkpoint with the best macro {\em F\textsubscript{1}} score on the development sets to evaluate performance on the respective test sets. To limit GPU usage during our  experiments, we normalize all datasets considered  by limiting the size of Train, Dev, and Test splits to \textit{50k, 5k, 5k } samples respectively.\footnote{For example, for the \textit{Age} and \textit{Gender} datasets, Train, Dev, and Test  splits have $1.3$m, $160$k, and $160$k, respectively. So, we randomly pick $50$k, $5$k, and $5$k samples respectively.}

\section{Results}

As explained, we compare different methods on $15$ different Arabic social media datasets involving binary and multiclass classification. We present performance of the methods in Table \ref{table:model_performance}. Evidently, CL-based methods achieve better performance on majority of the tasks. On average, three out of five CL-based methods (LCL, SCL, and TACT) achieve better performance than CE-MARBERT. Overall, LCL achieves the best F\textsubscript{1}-score averaging across all the tasks.

It is important to note that there is no unique superior method across the tasks. This shows that CL-based methods can be task-specific, depending on the nature of how they are formulated. For example, LCL performs well on multiclass datasets such as \textit{Abusive} and \textit{AraNeT\textsubscript{emo}}, while TLCL performs well on \textit{SemEval\textsubscript{emo}}. LCL and TLCL adopt more fine-grained representations with the incorporation of the weighting network which consequently helps them distinguish confused classes. However, for \textit{Dialect at RegionLevel}, we speculate that since the labels are already fine-grained, it is more important to improve the robustness rather than inter-label relationship. Therefore, CAT achieves best performance on this task, followed by TLCL. Similarly, on binary classification tasks such as \textit{hate speech} and \textit{Offensive language detection}, where a subtle semantic change in meaning can alter the labels, robust methods are expected to outperform others. Therefore, adversarial methods like CAT and TACT achieve better F\textsubscript{1}-score.

For most of the tasks, F\textsubscript{1}-scores obtained from different CL-methods are close to each other and the vanilla SCL achieves similar average score to the other models. This proves that although task-specific formulation may help the models to improve on a certain task, the most important factor evolves around the fundamental \textit{minmax} nature of contrastive learning which is minimizing the distance among the representations of the same class while maximizing the distance among the representations of the different classes.

\section{Analysis}

\subsection{Data Efficiency}

\begin{table*}[t]
\centering
\small 
\setlength{\tabcolsep}{6pt}
\begin{tabular}{lcccclcccclcccc}
\hline
     & \multicolumn{4}{c}{{\bf Dialect-Country}}                                                                                                                          &  & \multicolumn{4}{c}{{\bf Dialect-Region}}                                                                                                                      &  & \multicolumn{4}{c}{{\bf AraNeT\textsubscript{emo}}}                                                                                                                                 \\ \hline
     & \multicolumn{1}{r}{\textbf{10\%}}      & \multicolumn{1}{r}{\textbf{25\%}}      & \multicolumn{1}{r}{\textbf{50\%}} & \multicolumn{1}{r}{\textbf{100\%}}     &  & \multicolumn{1}{r}{\textbf{10\%}}      & \multicolumn{1}{r}{\textbf{25\%}} & \multicolumn{1}{r}{\textbf{50\%}} & \multicolumn{1}{r}{\textbf{100\%}}     &  & \multicolumn{1}{r}{\textbf{10\%}}      & \multicolumn{1}{r}{\textbf{25\%}}     & \multicolumn{1}{r}{\textbf{50\%}} & \multicolumn{1}{r}{\textbf{100\%}}     \\ \cline{2-5} \cline{7-10} \cline{12-15} 
CE   & 27.78                                  & 30.5                                   & 30.91                             & 32.84           &  & 63.09                                  & 63.16                             & 63.59                             & 65.29           &  & 53.85                                  & 56.73                                 & 59.18                             & 62.47           \\
SCL  & 28.49                                  & \textbf{31.87} & \textbf{32.89}                    & \textbf{33.63} &  & 63.08                                  & 63.23                             & 63.37                             & 64.78                                  &  & 54.47                                  & 58.35                                 & 58.35                             & 61.49                                  \\
CAT  & 26.57                                  & 30.33                                  & 32.71     & 33.24                                  &  & \textbf{64.32} & \textbf{65.3}                     & \textbf{65.42}                    & \textbf{65.54} &  & 54.75                                  & 54.03                                 & 55.51                             & 59.31                                  \\
TACT & 27.63                                  & 29.88                                  & 32.04                             & 32.69                                  &  & 63.8                                   & 64.1      & 64.32     & 64.56                                  &  & 53.27                                  & \textbf{59.3} & 59.18                             & 57.99                                  \\
LCL  & \textbf{28.97} & 30.5                                   & 31.78                             & 33.62           &  & 63.72                                  & 64.72                             & 65.06                             & 62.92           &  & \textbf{55.47} & 59.25                                 & 62.21     & 62.56           \\
TLCL & 27.69                                  & 30.44                                  & 32.18                             & 31.34                                  &  & 62.71                                  & 64.53                             & 64.6                              & 64.92                                  &  & 54.62                                  & 59.31                                 & \textbf{62.98}                    & \textbf{64.13} \\ \hline
\end{tabular}
\caption{
\label{table:data_efficiency}
Model performance on varying dataset sizes. \textbf{Bold} values represent the best performance for a particular dataset and dataset size.
}
\end{table*}

To investigate how the methods perform with limited data, we train the models under different size constraints using three datasets (one binary and two multiclass). We present results of this set of experiments in Table \ref{table:data_efficiency}. One interesting observation is that improvement in performance is not always monotonic with respect to data size. We believe that larger-sized training sets only aid models with test samples with idiosyncrasies and that small training sets sufficiently cover a wide range of data distributions.
However, we observe that CE-MARBERT fails to outperform CL-based methods in any constraint. Specifically, for \textit{Dialect at CountryLevel} dataset, $50\%$ of the data is sufficient for SCL to outperform CE-MARBERT trained on the full dataset. Additionally, CAT achieves comparable performance to CE-MARBERT with $50\%$ training data. For \textit{Dialect at RegionLevel} dataset, only $10\%$ training data is sufficient for CAT, TACT, and LCL to outperform CE-MARBERT with $50\%$ training data. Moreover, CAT requires only $50\%$ training data to outperform CE-MARBERT with full training data. Finally, for \textit{AraNeT\textsubscript{emo}} dataset, LCL, TACT, and TLCL with $25\%$ training data outperform CE-MARBERT with $50\%$ training data. TLCL with $50\%$ data outperforms CE-MARBERT with full (i.e., $100\%$) training data while LCL with $50\%$ data achieves similar performance. \textit{This analysis shows that enhancing the representations of different classes via CL helps the model to produce more distinguishable clusters. As a result, the models require only smaller training data to project a sample to a particular class.}

\subsection{Impact of Batch Size}

\begin{figure}[h]
     \centering
     \begin{subfigure}[b]{0.3\textwidth}
         \centering
         \includegraphics[width=\textwidth]{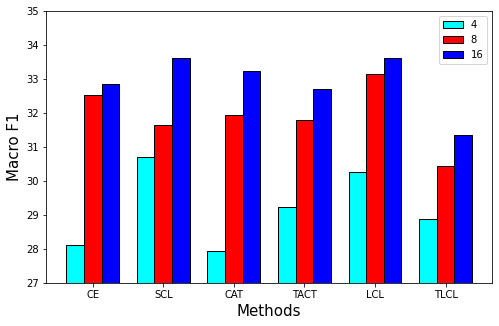}
         \caption{Dialect at CountryLevel}
     \end{subfigure}
     \hfill
     \begin{subfigure}[b]{0.3\textwidth}
         \centering
         \includegraphics[width=\textwidth]{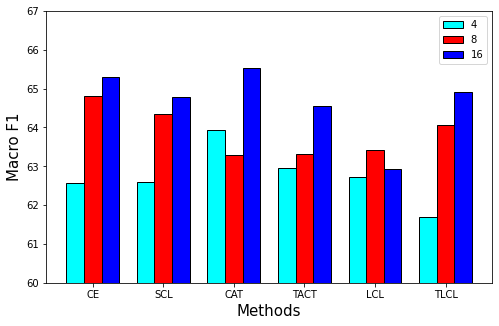}
         \caption{Dialect at RegionLevel}
     \end{subfigure}
     \hfill
     \begin{subfigure}[b]{0.3\textwidth}
         \centering
         \includegraphics[width=\textwidth]{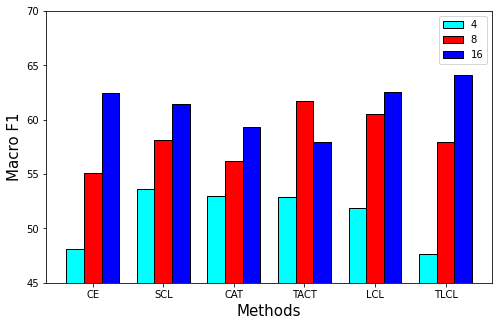}
         \caption{AraNeT\textsubscript{emo}}
     \end{subfigure}
        \caption{
        \label{fig:batch_size}
        Ablation study on the impact of batch size on performance of the models.
        }
\end{figure}


We study how batch size affects model performance. We consider batch sizes of $4, 8, 16$ on three datasets, showing performance in Figure~\ref{fig:batch_size}.
We observe that, with only a few exceptions, performance of the models increases along with the increase of batch size. Larger batch sizes contain more samples from different classes, which helps the model to learn better via comparing these samples. Our analysis corroborates findings of prior works such as~\citet{moco},~\citet{mocose}, 
and~\citet{carl} that propose the incorporation of a separate memory bank to hold the negative samples for comparison.

\subsection{Visualization of Representations}

\begin{figure}[h]
     \centering
     \begin{subfigure}[b]{0.2\textwidth}
         \centering
         \includegraphics[width=\textwidth]{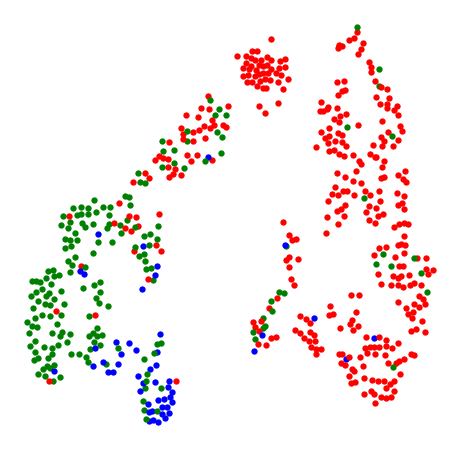}
         \caption{CE}
     \end{subfigure}
     \hfill
     \begin{subfigure}[b]{0.2\textwidth}
         \centering
         \includegraphics[width=\textwidth]{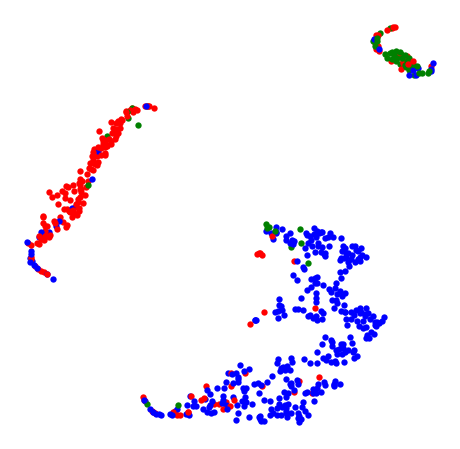}
         \caption{SCL}
     \end{subfigure}
     \hfill
     \begin{subfigure}[b]{0.2\textwidth}
         \centering
         \includegraphics[width=\textwidth]{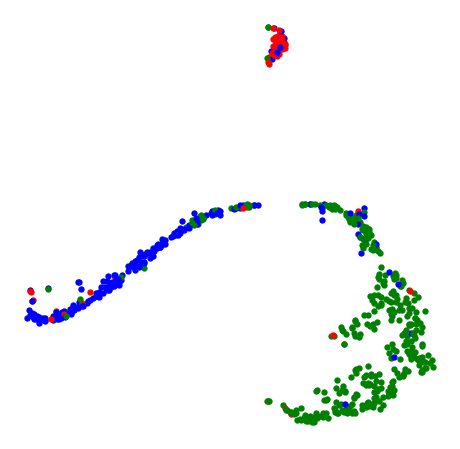}
         \caption{CAT}
     \end{subfigure}
    \hfill
     \begin{subfigure}[b]{0.2\textwidth}
         \centering
         \includegraphics[width=\textwidth]{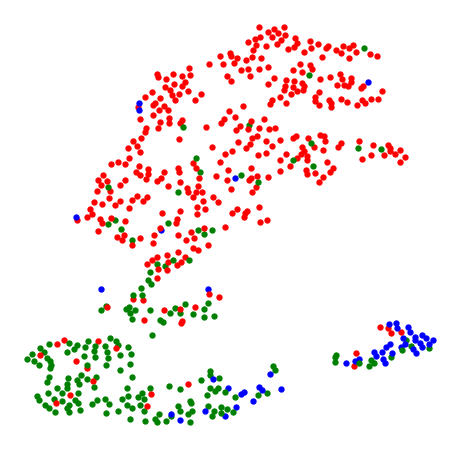}
         \caption{TACT}
     \end{subfigure}
    \hfill
     \begin{subfigure}[b]{0.2\textwidth}
         \centering
         \includegraphics[width=\textwidth]{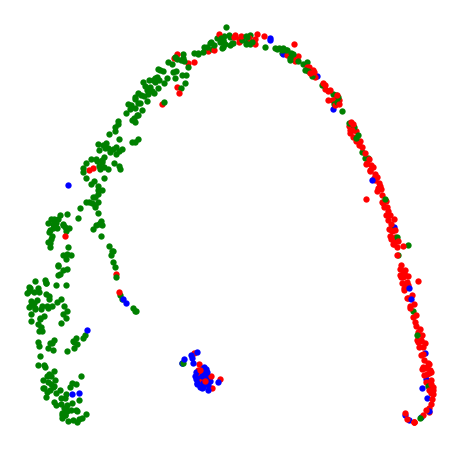}
         \caption{LCL}
     \end{subfigure}
    \hfill
     \begin{subfigure}[b]{0.2\textwidth}
         \centering
         \includegraphics[width=\textwidth]{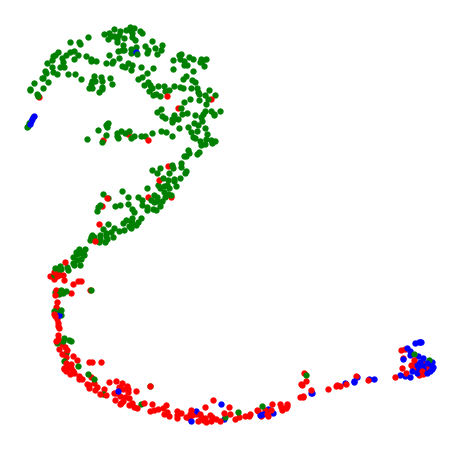}
         \caption{TLCL}
     \end{subfigure}
        \caption{t-SNE representations of the validation set of \textit{abusive} dataset (\textcolor{green}{green} = normal, \textcolor{red}{red} = abusive, \textcolor{blue}{blue} = hate).}
        \label{fig:tsne}
\end{figure}


We plot t-SNE representations of the test samples from the \textit{Abusive} dataset in Figure~\ref{fig:tsne}. The representations are colored with true labels. We notice that CL-based methods cluster \textcolor{green}{\textit{normal}} and \textcolor{red}{\textit{abusive}} samples far from each other, unlike CE-MARBERT. Since CL attempts to maximize the distance between different classes, it helps the models produce more distinct clusters. Additionally, LCL and TLCL methods cluster \textcolor{red}{\textit{abusive}} and \textcolor{blue}{\textit{hate}} classes better than other methods. Since, they capture inter-label relations, the methods identify confusable examples of \textcolor{red}{\textit{abusive}} and \textcolor{blue}{\textit{hate}} better than other methods.

\section{Limitations}

An inherent limitation of CL methods is their reliance on hyperparameters. In particular, they are sensitive to batch size. Larger batch sizes usually yield better performance. Other hyperparameters like $\tau$ and $\lambda$ can also impact performance given a specific task. Lastly, the accommodation of larger batch size comes at the cost of higher computational resources.  

\section{Conclusion}

In this work, we study various supervised contrastive learning methods for a wide range of Arabic social meaning tasks. We show that CL-based methods outperform generic cross entropy finetuning for majority of the tasks. Through empirical investigations, we find that improvements resulting from applying CL methods are task-specific. We interpret these results vis-a-vis different downstream tasks, with a special attention to the number of classes involved in each task. Finally, we demonstrate that CL methods can achieve better performance with limited training data and hence can be employed for low-resource settings. 

In the future, we plan to extend our work beyond sentence classification by experimenting on tasks such as  token-classification and question-answering. Our work stands as a comprehensive investigation of applying contrastive learning to Arabic social meaning. We hope this work will trigger further investigations of CL in Arabic NLP in general.

\bibliography{anthology,custom,lrec2020_araNet}
\bibliographystyle{acl_natbib}




\end{document}